\documentclass[letterpaper, 10 pt, conference]{ieeeconf}  % Comment this line out
\IEEEoverridecommandlockouts                              % This command is only
\usepackage[utf8]{inputenc}
\usepackage[T1]{fontenc}

\usepackage{cite}
\usepackage{amsmath,amssymb,amsfonts}
\usepackage{graphicx}% http://ctan.org/pkg/graphicx
\usepackage{caption}% http://ctan.org/pkg/caption
\usepackage{lipsum}% http://ctan.org/pkg/lipsum
\usepackage{graphicx}
\usepackage{adjustbox}
\usepackage{algpseudocode}
\usepackage[]{algorithm2e}
\usepackage{amsmath,amssymb}

\newcommand{\Xspace}        {{\mathbb X}}

\title{\LARGE \bf
TDA-Net: Fusion of Persistent Homology and Deep Learning Features for COVID-19 Detection From Chest X-Ray Images
}

\author{Mustafa Hajij*$^{1}$, Ghada Zamzmi* $^{2}$, and Fawwaz Batayneh$^{3}$% <-this % stops a space
\thanks{$^{1}$M. Hajij is with Faculty of Santa Clara University, Santa Clara, CA, USA, joint first author.}%
\thanks{$^{2}$G. Zamzmi is with the Department of Computer Science and Engineering, University of South Florida, FL, USA, joint first author.}
\thanks{$^{3}$F. Batayneh is with the University of Queensland, Brisbane, Australia.}
}

\begin{document}

\maketitle
\thispagestyle{empty}
\pagestyle{empty}

%%%%%%%%%%%%%%%%%%%%%%%%%%%%%%%%%%%%%%%%%%%%%%%%%%%%%%%%%%%%%%%%%%%%%%%%%%%%%%%%
\begin{abstract}
Topological Data Analysis (TDA) has emerged recently as a robust tool to extract and compare the structure of datasets. TDA identifies features in data (e.g., connected components and holes) and assigns a quantitative measure to these features. Several studies reported that topological features extracted by TDA tools provide unique information about the data, discover new insights, and determine which feature is more related to the outcome. On the other hand, the overwhelming success of deep neural networks in learning patterns and relationships has been proven on various data applications including images. To capture the characteristics of both worlds, we propose \textit{TDA-Net}, a novel ensemble network that fuses topological and deep features for the purpose of enhancing model generalizability and accuracy. We apply the proposed \textit{TDA-Net} to a critical application, which is the automated detection of COVID-19 from CXR images. Experimental results showed that the proposed network achieved excellent performance and suggested the applicability of our method in practice. 

\end{abstract}

%%%%%%%%%%%%%%%%%%%%%%%%%%%%%%%%%%%%%%%%%%%%%%%%%%%%%%%%%%%%%%%%%%%%%%%%%%%%%%%%
\section{INTRODUCTION}
Deeply concerned by the alarming levels of spread and severity, the World Health Organization (WHO) declared COVID-19 as an ongoing pandemic in March 11, 2020. Since then, there have been more than 195 millions confirmed cases and over four millions reported deaths worldwide \cite{cdc}. The real time reverse transcriptase polymerase chain reaction (RT-PCR) remains the current standard for diagnosing COVID-19 disease \cite{udugama2020diagnosing}. Other diagnostic tests for COVID-19 include the loop-mediated isothermal amplification (LAMP), the lateral flow, and enzyme-linked immunosorbent assay (ELISA) \cite{udugama2020diagnosing}. 

In addition to these techniques, imaging techniques, such as Chest X-ray (CXR) and computed tomography (CT), have been used for early COVID-19 diagnosis and prediction of disease progression \cite{shi2020review}. While CT scans have shown higher sensitivity in detecting pneumonia infection and pulmonary manifestations \cite{shi2020review} as compared to CXR, CT imaging techniques is not widely used for COVID-19 diagnosis due to several issues including the high cost of CT, non-portability, and cross-contamination issues \cite{shi2020review}. CXR imaging technique provides a cheaper alternative that facilitates the diagnosis of patients who cannot move, and hence reduces cross-contamination. Since the pandemic started, the shortage of expert radiologists has been increasing especially in third-world countries \cite{davarpanah2020novel}. To mitigate this shortage and speedup the diagnosis of COVID-19, artificial intelligence (AI) driven computer-aided diagnostic (CADx) tools can be used for Point-of-Care Testing (POCT).

Deep learning provides powerful tools for extracting features that carry rich texture information about the images. Since early this year, Covid-19 has attracted much attention from the deep learning community. Numerous works (e.g., \cite{bai2020performance,rajaraman2020iteratively,wu2020jcs,li2020artificial}) propose to use convolutional neural network (CNN) for detecting COVID-19 viral pneumonia from CXR \cite{ozturk2020automated,rajaraman2020iteratively} and CT images \cite{wu2020jcs,bai2020performance,li2020artificial}. On the other hand, TDA \cite{edelsbrunner2010computational,carlsson2009topology} has  emerged  recently  as  a  robust  tool  to extract and compare the structure of datasets. TDA identifies specific features (e.g., connected components and holes) and describes the extent to which they persist across the image. Several studies reported that \cite{nielson2015topological,joshi2019survey} topological features extracted by TDA tools (e.g., \textit{persistent homology} \cite{edelsbrunner2008persistent} and \textit{Mapper} \cite{singh2007topological}) provide unique information about the data, and determine which predictors are more related to the outcome. These characteristics allow TDA to provide better explainability as compared to deep learning methods. In addition, the topological analysis of the image allows to extract features that are invariant to the spatial transformation and more robust to noise \cite{zheng2015application,bae2017beyond}. TDA-based methods have shown excellent performance in several applications including neuroscience~\cite{ DabaghianMemoliFrank2012, giusti2016two, LeeKangChung2012b}, bioscience~\cite{dewoskin2010applications,hajij2020graph},  and images \cite{clough2019topological,garside2019topological}, among others.

This work is the first that utilizes the quantitative power available in TDA tools and applies it to detect COVID-19 from CXR images. The main contribution of this paper can be summarized as follows:
\begin{itemize}
    \item We propose TDA-Net, a novel network that fuses topological and deep features. TDA-Net contains two branches: a deep branch that takes a raw image as input and a topological branch that takes a topological feature vector as input. The outputs of both branches are then fused and used to perform classification. The deep branch provides rich texture information while the topological branch provides rich representation of the data shape, and extract invariant and robust features. To the best of our knowledge, we are the first to fuse topological and deep learning features extracted from medical images. 

    \item We propose to apply our proposed network to a critical application, which is the automated detection of COVID-19 from CXR images. The experimental results with two publicly available datasets showed the effectiveness of the proposed network as compared to baseline models. 
\end{itemize}

\section{Background}
In this section, we provide technical background and explain the process of converting a given tensor (i.e., a multidimensional array such as a 2d image), to a vectorized representation of its persistence diagram.

\subsection{Homology}
\label{1}

Homology deals with the topological features of a space.  
Given a topological space $\Xspace$,
the $0$-, $1$-, and $2$-dimensional homology groups, denoted as 
$H_0(\Xspace)$, 
$H_1(\Xspace)$, and $H_2(\Xspace)$ respectively, correspond to 
(connected) components, tunnels, and voids of $\Xspace$.
In general, the $k$-th \emph{homology} group $H_k(\Xspace)$ describes 
the $k$-dimensional holes in $\Xspace$; the $k$-th \emph{Betti number} $\beta_k$ is the rank of this group, that is, $\beta_0$ is the number of components, $\beta_1$ is the number of $1$-dimensional (circular) holes/tunnels, and $\beta_2$ is the number of $2$-dimensional holes/voids.

\begin{figure}[h]
    %\vspace{-5pt}
	\centering
	{\includegraphics[width=0.99\linewidth]{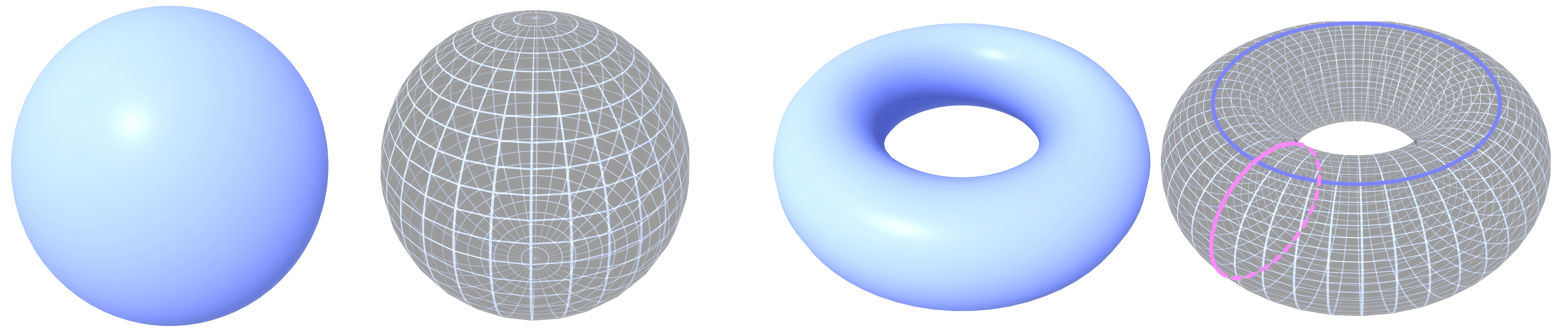}}
	\caption{Betti numbers for a sphere and torus.}
	\label{fig:betti}
    %\vspace{-5pt}
\end{figure}

For example, a circle contains a single component and a $1$-dimensional tunnel, but no higher-dimensional holes---$\beta_0 = 1$, $\beta_1 = 1$ and $\beta_i = 0$ ($i > 1$); a sphere (Figure~\ref{fig:betti}) contains a single component and a $2$-dimensional void, but no $1$-dimensional tunnel---$\beta_0 = 1$, $\beta_1 = 0$, $\beta_2 = 1$ and $\beta_i = 0$ ($i >2$); and a torus (Figure~\ref{fig:betti}) contains a single component, 2 tunnels (Figure~\ref{fig:betti} (bottom) one in blue and one in pink), and a void---$\beta_0 = 1$, $\beta_1 = 2$, $\beta_2 = 1$ and $\beta_i = 0$ ($i >2$).

\subsection{Persistent Homology on Tensors}
\label{3}
While homology is applicable on simplicial complexes, it is not immediately applicable to real-world data such as images, and hence, persistent homology comes into play. Essentially, persistent Homology quantifies the same topological information in data by transforming them into a filtration (a nested sequence of spaces), and then performing persistent homology computation on the filtration followed by storing the final result into a multi-set structure in $\mathbb{R}^2$ called the persistence diagram. Intuitively, a filtration in our context is the tool to extract features from the tensors and the persistence diagram is the data structure that stores these features. 

On the other hand, the persistence diagram can be thought of as a function that takes a \textit{filtration} as an input and produces a multi-set of points $\mathbb{R}^2$. Precisely, let $K$ be a simplicial complex and $V(K)$ the vertices of $K$. Let $S$ be an ordered sequence $\sigma_1,\cdots,\sigma_n $ of all simplices in $K$, such that for simplex $\sigma \in K $ every face of $\sigma$ appear before it $\sigma$ in $S$. Then, $S$ induces a nested sequence of subcomplexes called a \textit{filtration}:
$\phi=K_0  \subset K_1 \subset ... \subset K_n = K$.
 A $d$-homology class $\alpha \in H_d(K_i)$ is said to be \textit{born} at the time $i$ if it appears for the first time as a homology class in $H_d(K_i)$. A class $\alpha$ \textit{dies} at time $j$ if it is trivial $H_d(K_j)$ but not trivial in $H_d(K_{j-1})$. The \textit{persistence} of $\alpha$ is defined to be $j-i$. Persistent homology captures the birth and death events in a given filtration and summarizes them in a multi-set structure called the \textit{persistence diagram} $P^d(\phi)$. Specifically, for any integer $d\geq0$, the $d$-persistence diagram of the filtration $\phi$ is a collection of pairs $(i,j)$ in the plane where each $(i,j)$ indicates a $d$-homology class that is created at time $i$ in the filtration $\phi$ and killed entering time $j$.  
 
 Persistent homology can be defined given \textit{any} filtration. For the purpose of this work, we assume that the input is a 2-d tensor and utilize this tensor to build a \textit{lower-star filtration} which reflects sublevel topology of this tensor. We illustrate next how to convert a tensor to a simplicial complex.
 
 From a tensor $M$ specified as an 2-d array, we build a 1-d simplicial complex $K(M)=(V(M),E(M))$ as follows. Every pixel in the tensor is a vertex $v$ in the set of vertices $V(M)$, and every vertex $v$ in $V(M)$ is connected to the set of neighbor pixels in $M$ that are immediately adjacent to $v$. In our case, we connect every pixel to the $8$ other vertices obtained by considering the $8$ immediate neighbors of the pixel that this vertex represents. Clearly, at the boundary of a tensor, every vertex is connected to fewer vertices. See Figure 2 for an example of converting 2d-tensor $M$ to the complex $K(M)$ \footnote{ This method can be generalized to  n-d multidimensional tensors. Moreover, there are multiple ways to build a complex out of a tensor. Typically, in the context of images, cubical complexes are popular because they are more efficient  \cite{allili2001cubical}. For our particular case study, the difference in complex size is negligible.  }. 

\begin{figure}[h]
    %\vspace{-5pt}
	\centering
	{\includegraphics[width=0.85\linewidth]{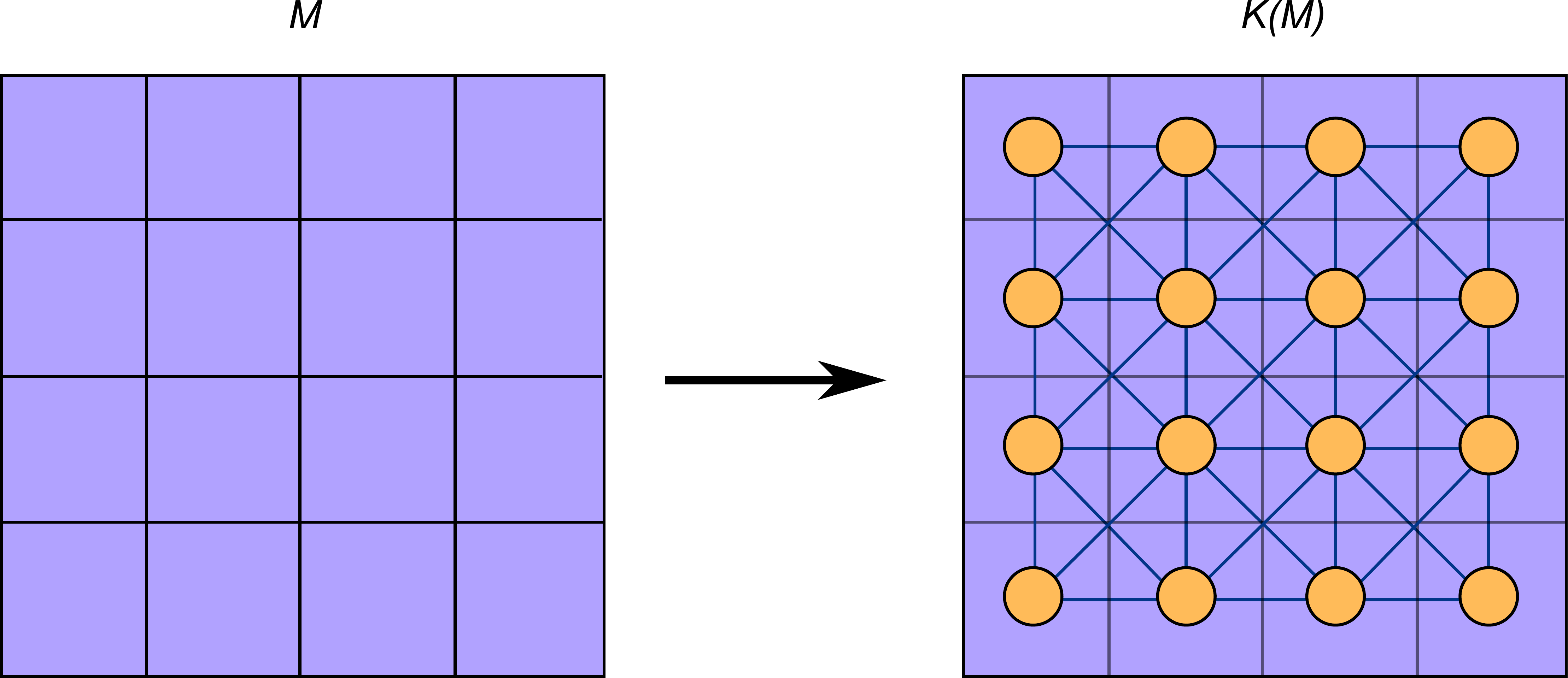}}
	\caption{Example of converting an image $M$ to a 1-d simplicial complex $K(M)$.   }
	\label{fig:SC}
    %\vspace{-5pt}
\end{figure}

 Using the above construction, we can think about a single channel image $M$ as a piecewise linear function $f_M:K(M)\to \mathbb{R}$ defined on the vertices of complex $K(M)$ where the value of $f_M$ at the a certain vertex is simply the value of the corresponding pixel. We extend $f$ to each edge of $K(M)$  by considering the value of that edge to be the maximum of the two pixel values it connects. Now, let $V(M)=\{v_1,\cdots,v_n\}$ be the set of vertices of $K(M)$ sorted in non-decreasing order of their $f$-values, and let $K_i(M):= \{\sigma \in K(M) | \max_{v \in \sigma}f_M(v)\leq f_M(v_i)   \} $. 
The lower-star filtration of $K(M)$ with respect to $f$, denoted by $\mathcal{F}_f(K(M))$, is defined as: 
\begin{equation}
\label{filter2}
  \phi=K_0(M)  \subset K_1(M) \subset ... \subset K_n(M) = K(M).
\end{equation} 

The lower-star filtration reflects the topology of the function $f_M$ in the sense that the persistence homology induced by filtration \ref{filter2} is identical to the persistent homology of the sublevel sets of the function $f_M$. 

In this work, we only consider the $0$-dimensional persistence diagram of the lower-star filtration of $M$, which we denote by $PD_0(M)$. Figure \ref{fig:SC} shows an example of computing the 0-persistence diagram of a 1-d tensor.

\begin{figure}[!h]
    %\vspace{-5pt}
	\centering
	{\includegraphics[width=0.95\linewidth]{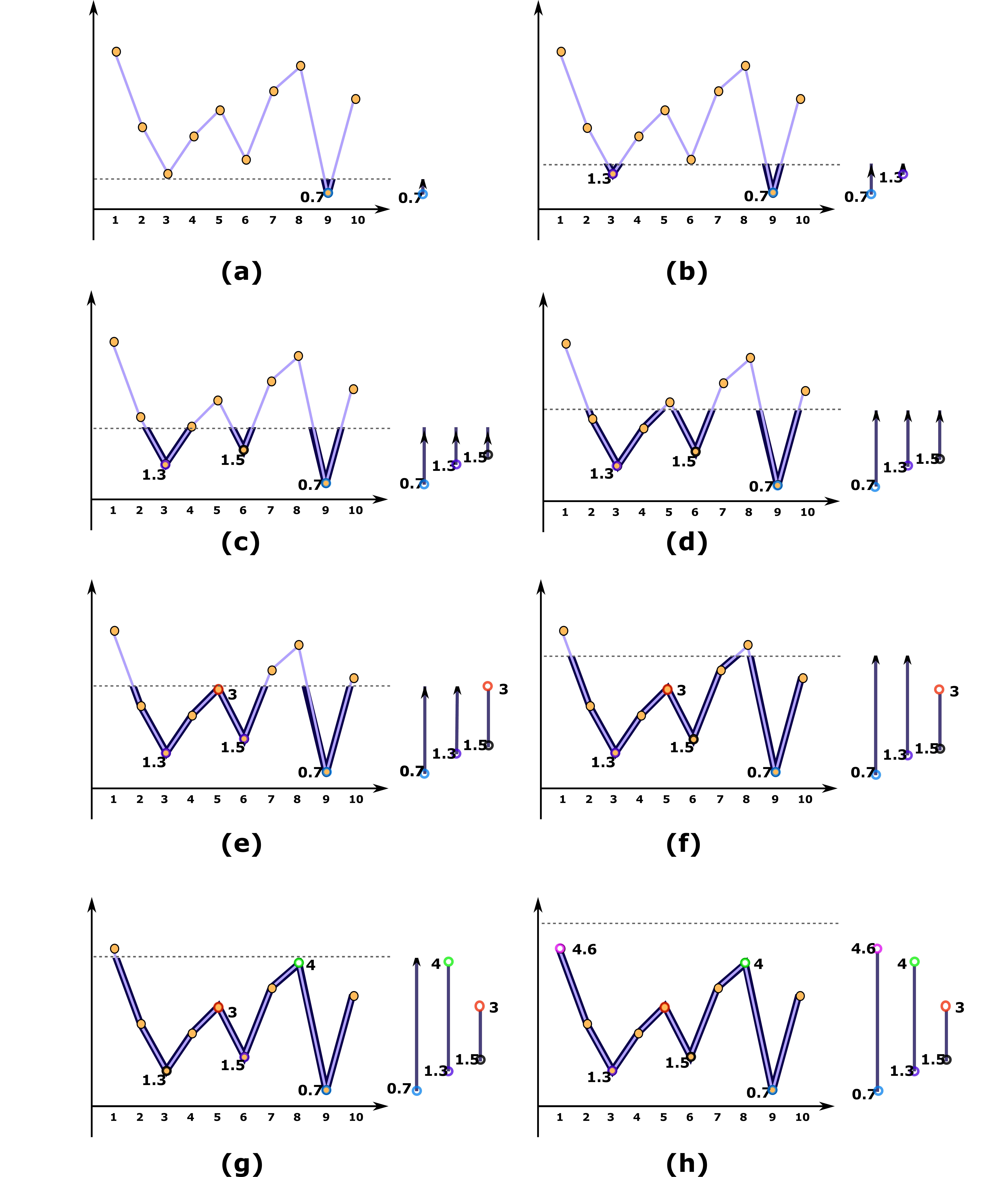}}
	\caption{An example of computing the persistence diagram of 1-d tensor. The persistence diagram consists of a collection of bars, each bar represents a topological feature and is encoded by a pair $(u, v) \in \mathbb{R}^2$ which represents the birth and the death of topological feature. The persistence of $(u,v)$ is simply given by $|v -u|$. Features with higher persistence values carry more significant topological information while features with low persistence are considered noise. In this example, the persistence diagram consists of three features : $(0.7,4.6)$, $(1.3,4)$ and $(1.5,3)$. }
	\label{fig:SC}
    %\vspace{-5pt}
\end{figure}

\subsection{Vector Representation of Persistence Diagram}

\begin{figure}[!h]
    %\vspace{-5pt}
	\centering
	{\includegraphics[width=0.99\linewidth]{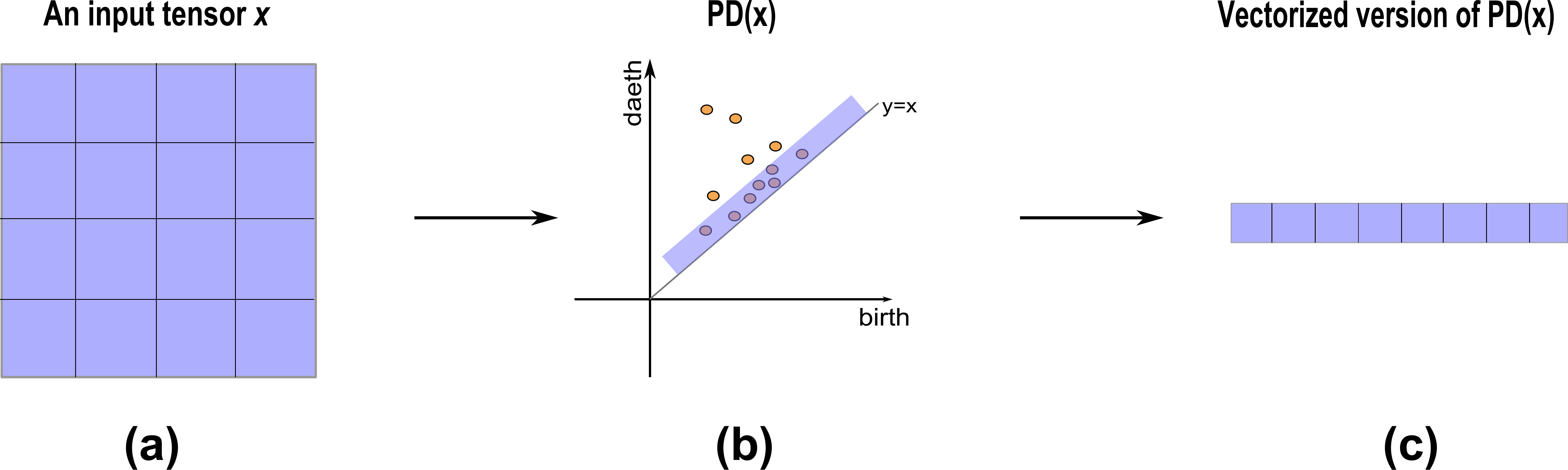}}
	\caption{The process of converting an input tensor to a vectorized version of the persistence diagram.  }
	\label{fig:network123}
    %\vspace{-5pt}
\end{figure}

Recently, there have been many attempts to \textit{vectorize} the persistence diagram in order to utilize it within a traditional machine learning framework. Generally speaking, the vectorization framework of the persistence diagram, illustrated in Figure \ref{fig:network123}, starts with a set of data of interest, say a set of images. Each element in this set is then converted to a persistence-based representation. By viewing these persistence representations as points in a feature space, we can define an appropriate distance or a kernel inside this feature space followed by performing a data analysis task. 

Many such vectorization schemes have been suggested recently. This includes betti curve \cite{umeda2017time}, the persistence landscape \cite{bubenik2015statistical}, the persistence image \cite{adams2017persistence} among others \cite{chen2015statistical,berry2020functional,kusano2016persistence}. While any of these tensors can be utilized, we choose to utilize the betti curves \cite{umeda2017time} representation of the persistence diagram.

%In a TDA-Net architecture, one may consider utilizing any of these tensors into the topological stream. In t,  we choose to work with the betti curves \cite{umeda2017time} represeantation of the persistence diagram.

%From a deep learning setting, these objects are essentially tensors that can be utilized into a deep net for a given learning task. For instance a persistence image, in simple terms, is 2-D tensor manifestation of the PD. On the other hand the betti curve and the persistence landscape are 1-D tensor of PD.

\section{Approach}

\subsection{TDA-Net}
\begin{figure}[h]
    %\vspace{-5pt}
	\centering
	{\includegraphics[width=\linewidth]{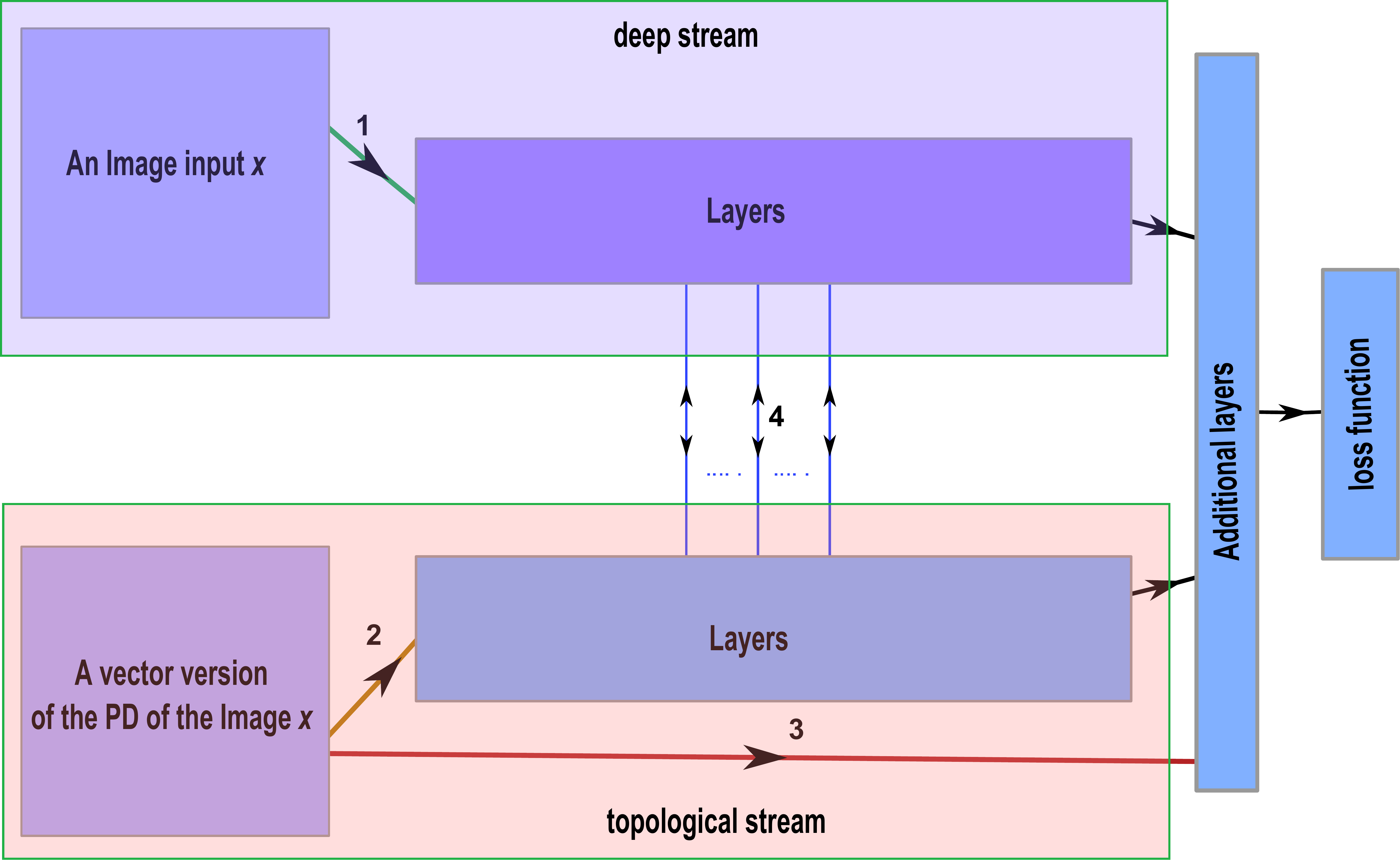}}
	\caption{An abstract sketch of the general purpose TDA-Net. The input for the model consists of the raw image data and a vector version of the persistence diagram (PD). The PD can be injected into the model as an input along with the raw image data (as indicated in the yellow arrow) or concatenated to a later activation (as indicated in the red arrow) or both.}
	\label{fig:network}
    %\vspace{-5pt}
\end{figure}

As shown in figure \ref{fig:network}, the proposed TDA-Net consists of two sub-neural networks with two different data inputs: the raw pixel image, denoted by $x$, and the vector version of the persistence diagram of the input image $x$. The sub-network of TDA-Net that takes the raw data is called the \textit{deep stream} and the sub-network that takes the persistence diagram is called the \textit{topological stream}. Note that Figure \ref{fig:network} illustrates an abstraction for a general purpose TDA-net and the design of the network admits multiple variations. In particular, the persistence diagram can be injected into the topological stream immediately as an input (as indicated in the yellow arrow labeled by "2" in 
Figure \ref{fig:network}) or concatenated to a later activation (as indicated by the red arrow labeled by "3" in Figure \ref{fig:network}) or both. Moreover, one can add various connections between the two streams. In this work, we explore multiple versions of TDA-Nets as presented in Section \ref{tests}.

%\subsubsection{Why TDA-Nets ?} There are multiple advantages of a TDA-Net architecture. On one hand, TDA-Nets can be thought of as ensemble models where the topological stream and the deep stream form the simple models that are combined together to form a  more sophisticated model. Ensemble model have proven to be very effective in practice and have been applied to medical images applications \cite{iqbal2019deep,kamnitsas2017ensembles,hijazi2019ensemble}. On the other hand, Deep learning and persistent homology share common elements that make them both effective and robust for vast variety of tasks. For instance, neural networks have been proven to provide better data representation than the raw data, especially in the context of CNNs. Moreover, encodings of data induced from persistent homology, i.e., the persistence diagram, seem to be tracking different information from the encodings obtained in a neural network setting. This will be evident in the experiments that we conduct in section \ref{tests}. This suggests that combining the encodings obtained from persistent homology and neural network encodings could increase the quality of the learning task at hand.

\section{Experiments and Results}
\label{tests}
\begin{figure}[h]
    %\vspace{-5pt}
	\centering
	{\includegraphics[width=0.95\linewidth]{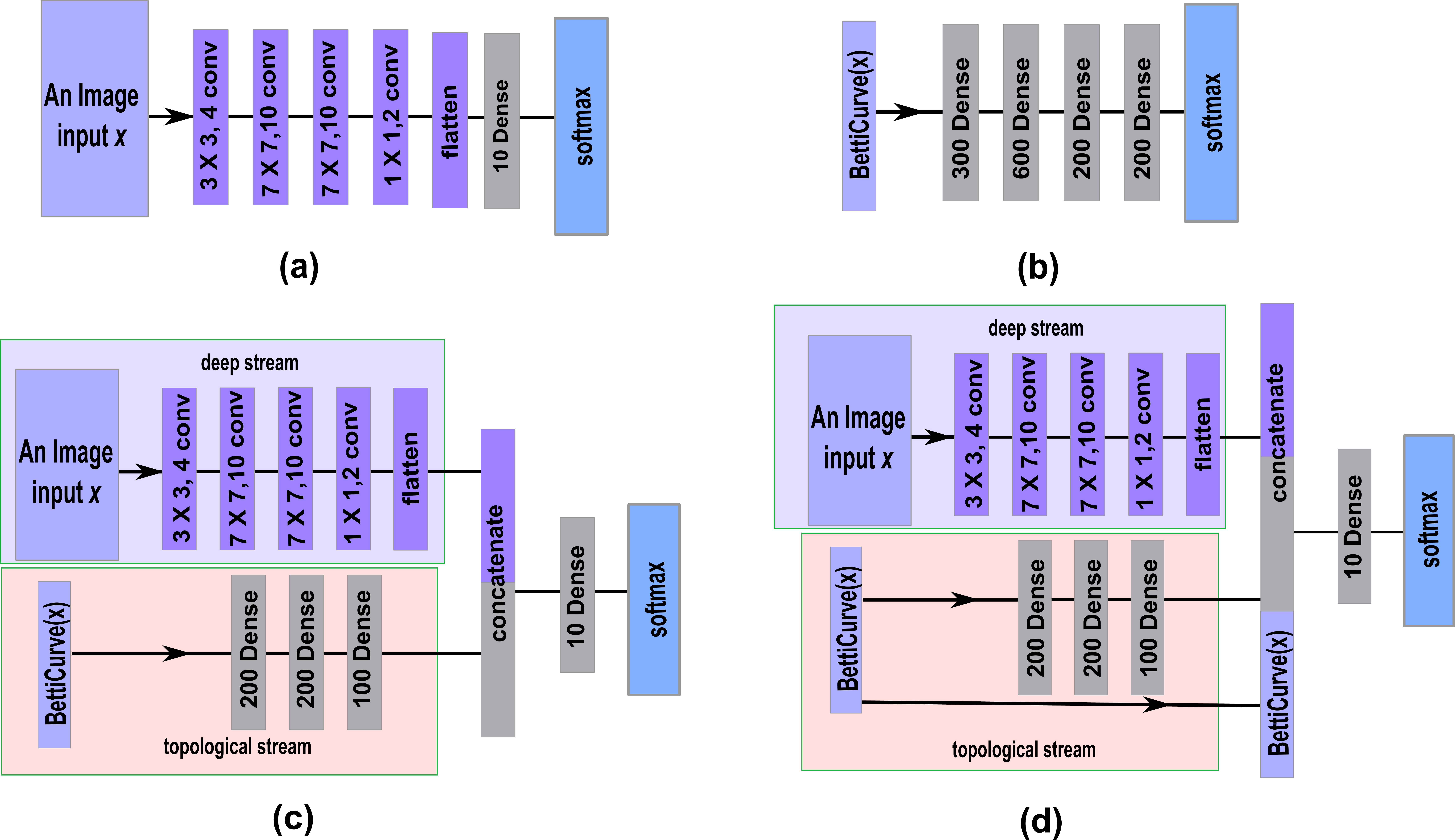}}
	\caption{The four neural networks utilized in this work. These nets are (a) the base model, (b) the $TDA-Net_{1}$ model that takes only the betti curve as input, (c) the $TDA-Net_{1,2}$ model that takes the raw image data and betti curve as inputs and finally (d) the $TDA-Net_{1,2,3}$ model that takes the betti curve of an image and the raw image data as inputs but also injects the betti curve into a later activation layer.}
	\label{fig:networks1234}
    %\vspace{-5pt}
\end{figure}

In this section, we evaluate 3 different versions of TDA-Nets against a simple base CNN model. Figure \ref{fig:networks1234} presents all the models used in this work.

We choose all the models to be simple with a small number of parameters for three reasons. First, simple and smaller models are efficient and more suitable for analysis in embedded real-time systems. Second, the complexity of the model makes it hard to analyze which features are more useful and effective for the learning application at hand. Finally, TDA-Net can be considered as a type of an ensemble model where a simple deep stream and a simple topological stream are combined together to enhance generalizability and performance.

% While the classification results indicated for the base model can probably be improved by utilizing different and more sophisticated archetype such as \cite{}. We did not choose this route for multiple reasons. On one hand, as we mention earlier, we like our models to execute quickly, preferably in real time on small devices. On the other hand, the complexity of the model makes it hard to analyse which features are more useful for the application at hand. 

%reached here

\subsection{Dataset \& Prepossessing}
We evaluate the proposed approach using two publicly available Chest X-ray datasets. The first dataset \cite{Cohenrepo} contains 351 chest X-ray images labeled as normal and COVID-positive (287 images). We only selected the COVID-positive images from this dataset. The other dataset is a Kaggle dataset \cite{kaggle} that has 112120 chest X-ray images labeled as COVID-negative (i.e., bacterial and non COVID viral pneumonia). To obtain a balanced dataset for evaluation, we used 287 images from this dataset as COVID-negative class. Thus, the final dataset has 287 COVID-negative images and 287 COVID-positive images. This dataset was divided into a training set and a leave-Out set. The leave-Out set consists of $\approx 20\%$ of the total dataset, which is 116 samples divided equally between the positive and the negative classes. During training, all images were scaled to $128 \times 128$. 

In order to incorporate the data for the topological stream of TDA-Net, we compute the persistence diagrams of all images in the constructed data followed by computing the betti curve vector representations of these diagrams. We chose to embed the betti curve vector in $\mathbb{R}^{100}$ \footnote{We empirically found this length to be sufficient to store all topological features in our dataset.}. The pipeline of converting a dataset into its topological incarnation is described in Figure \ref{fig:SC}

%Given the nature of our data's distributions, we are currently considering additional pre-processing to extract only the X-ray component from each and rotate each picture to the same orientation.  Additionally, we perform data augmentation if our models suffer due to data imbalance to boost the number of negative examples.

%One form of data augmentation that we have taken advantage of is normalizing the size and resolution of each picture. The final layer in our model consist of average pooling operation to allow images of any size to be passed through the convolutions before the fully connected layers.

\subsection{Networks Design and Configuration}
Four networks were trained to test our method. The first one is the base neural network, which is a simple CNN model that consists of 4 convolutional layers and two dense layers. Specifically, the input  for the neural network is input layer $128 \times  128 \times 3$. This input then goes through $4$ convolutional layers of the sizes indicated in Figure \ref{fig:networks1234} part (a). The final layer is a softmax layer. In the second network, we use a TDA-net that only consists of a topological stream. For this network, we only use the Betti-curve of the image $x$ as an input and we do not use any raw data information. Since the Bettie curve is essentially a vector, we choose a fully connected neural network architecture on this type of data. The details of this network, which we call $TDA-Net_{1}$, is given in Figure \ref{fig:networks1234}. We call this network $TDA-Net_{1}$ because it only has a connection of type $1$ as indicated in Figure \ref{fig:network}. 

The third network is a TDA-Net that has both topological stream and deep stream. The architecture of the deep stream is identical to the architecture of the base network. However, the layers in the topological stream are chosen to be dense layers since the input (betti-curve) is essentially a vector. The design of this network is given in Figure \ref{fig:networks1234} part $(c)$. We denote this network by $TDA-Net_{1,2}$ because it has connections of type $1$ and $2$ as indicated in Figure \ref{fig:network}. The forth network is similar to the third one but it also injects the betti-curve to a later activation as indicated in the Figure \ref{fig:networks1234} part $(d)$. We denote this network by $TDA-Net_{1,2,3}$. All our networks have been trained using Keras \cite{chollet2015keras} with TensorFlow backend \cite{abadi2016tensorflow}. The computation of the persistence diagram and its vectorized version were performed using scikit-tda \cite{scikittda2019}. 

\subsection{Results}
We evaluated the different variations of TDA-Net using the validation set, and reported the results  Table \ref{table:TLalgebra}. From the table, we can observe that the model that considers topological features only (i.e., $TDA-Net_1$) achieved similar performance to the base CNN model which only uses the raw data. We can also observe that the ensemble TDA-nets models (i.e., $TDA-Net_{1,2}$ and $TDA-Net_{1,2,3}$) achieved the best results as compared to base model and $TDA-Net_1$. Further, while $TDA-Net_{1,2,3}$ achieved the highest accuracy, we can observe that model $TDA-Net_{1,2}$ achieved similar accuracy to $TDA-Net_{1,2,3}$ while maintaining lower specificity or true negative rate (TNR). These results are promising and suggest the superiority of using TDA-Net with two streams (deep and topological) as compared to TDA-Net with a single stream and the base CNN.  This is attributed to the fact that each stream provides unique information, which allows the extraction of better feature representation.

%As shown in the table, the prelimnary results are promising and To study the applicability of the proposed model above with our small dataset, we perform the typical metrics on the validation dataset. The initial results, if they can be improved, show promising and interesting numbers. The results are reported in Table \ref{table:TLalgebra}. The most notable result for us is that the model that considers topological features alone, model $TDA-Net_1$, performs very closely to the base CNN model which processes the raw data alone. On the other hand, ensemble TDA-nets models, $TDA-Net_{1,2}$ and $TDA-Net_{1,2,3}$ yielded the better results. Specifically, while $TDA-Net_{1,2,3}$ yield the highest accuracy, in this case, model $TDA-Net_{1,2}$ is preferred because it yields lower specificity/true negative rate (TNR) while maintaining very close accuracy.  

\begin{table}[t]
\centering
\begin{footnotesize}
  \centering
  \begin{adjustbox}{width=\columnwidth,center}
\begin{tabular}{|c|c|c|c|c|}
\hline
 \multicolumn{1}{|c|}{}&
 \multicolumn{1}{c||}{Base model} & 
    \multicolumn{1}{c||}{$TDA-Net_1$} & 
 \multicolumn{1}{c|}{$TDA-Net_{1,2}$}&
  \multicolumn{1}{c|}{$TDA-Net_{1,2,3}$}\\
\hline
Accuracy  &
0.87& 
0.89 &
0.92 &
0.93 \\
\hline
Precision &
0.84& 
0.84 &
0.95 &
0.88  \\
\hline
Recall &
0.87& 
0.87 &
0.85 &
0.95 \\
\hline
f-1 score &
0.86& 
0.86 &
0.9 &
0.92 \\
\hline
TNR &
0.89 & 
0.88 &
0.97 &
0.91 \\
\hline
\end{tabular}
\end{adjustbox}
\end{footnotesize}
\hspace{5pt}
  \caption{Performance of the baseline and different variations of TDA-Net on COVID Chest X-Ray images. 
  %The table describes $L^2$ error of the braid group relations reported after training the networks $f$ and $g$. 
  } 
 \label{table:TLalgebra}
\end{table}

%\subsection{Training}

%For training weighted cross entropy loss was utilized since the data is unbalanced. 

%The data we trained our models on is highly unbalanced. Most of the data consists of examples that are labeled negative, $1-a\%$ of the data, and only $a\%$ are labeled positive. Given the unbalance nature of the training data typical cross entropy loss did not generalize well. To encounter this limitation we utilize \textit{weighted cross entropy} given in the formula:

%\begin{equation}
%\label{loss2}
%    L(\hat{y}, y)=- \sum_{i=1}^m w^+ y_ilog\hat{y}_i + w^- (1-y)_ilog(1-\hat{y}_i), 
%\end{equation}
%where $w^+$ and $w^-$ are the %weight associated with the positive and negative classes respectively.

\section{Conclusion and Future Work}
In this work, we proposed TDA-Net, an ensemble deep learning network that fuses topological and deep learning features into one network for the purpose of enhancing generalizability and accuracy. The initial study on Covid-19 x-ray images is promising and suggests the applicability of the proposed method in practice.

The study herein has multiple limitations that should be addressed in future work. For example, a larger dataset must be selected to further establish the applicability of our method in a practical setting. Furthermore, the four models that we selected were trained from scratch on our dataset. A transfer learning comparison utilizing base classical CNN models such as VGG-16 \cite{simonyan2014very} would most likely improve the results for all models.

%\vspace{20}

%%%%%%%%%%%%%%%%%%%%%%%%%%%%%%%%%%%%%%%%%%%%%%%%%%%%%%%%%%%%%%%%%%%%%%%%%%%%%%%%

\bibliographystyle{abbrv}

\bibliography{refs}

%\begin{thebibliography}{99}

%\bibitem{c1} G. O. Young, ``Synthetic structure of industrial plastics (Book style with paper title and editor),'' 	in Plastics, 2nd ed. vol. 3, J. Peters, Ed.  New York: McGraw-Hill, 1964, pp. 15--64.
%\end{thebibliography}

\end{document}